\documentclass[letterpaper]{article} 
\usepackage{aaai24}  
\usepackage{times}  
\usepackage{helvet}  
\usepackage{courier}  
\usepackage[hyphens]{url}  
\usepackage{graphicx} 
\usepackage{natbib}  
\usepackage{caption} 
\usepackage{algorithm}
\usepackage{algorithmic}

\usepackage{newfloat}
\usepackage{listings}
\DeclareCaptionStyle{ruled}{labelfont=normalfont,labelsep=colon,strut=off} 
\floatstyle{ruled}
\newfloat{listing}{tb}{lst}{}
\floatname{listing}{Listing}
\usepackage{amsmath}
\usepackage{amssymb}
\usepackage{booktabs}
\usepackage{multirow}
\usepackage{multicol}
\usepackage{makecell}
\usepackage{utfsym}
\usepackage{cleveref}
\Crefname{table}{Table}{Table}
\Crefname{figure}{Figure}{Figure}
\Crefname{equation}{Equation}{Equation}

\title{Angle Robustness Unmanned Aerial Vehicle Navigation in GNSS-Denied Scenarios}
\author{
    Yuxin Wang\textsuperscript{\rm 1}, 
    Zunlei Feng\textsuperscript{\rm 1}, 
    Haofei Zhang\textsuperscript{\rm 1}, 
    Yang Gao\textsuperscript{\rm 1}, 
    Jie Lei\textsuperscript{\rm 2},
    Li Sun\textsuperscript{\rm 3}\thanks{Corresponding author.}, 
    Mingli Song\textsuperscript{\rm 1}
}
\affiliations{
    \textsuperscript{\rm 1}Zhejiang University\\
    \textsuperscript{\rm 2}Zhejiang University of Technology\\
    \textsuperscript{\rm 3}Ningbo Innnovation Center, Zhejiang University\\
    \{yuxinwang, zunleifeng, haofeizhang, roygao, lsun, brooksong\}@zju.edu.cn, jasonlei@zjut.edu.cn
}

\usepackage{bibentry}

\begin{document}

\maketitle

\begin{abstract}
Due to the inability to receive signals from the Global Navigation Satellite System (GNSS) in extreme conditions, achieving accurate and robust navigation for Unmanned Aerial Vehicles (UAVs) is a challenging task.
Recently emerged, vision-based navigation has been a promising and feasible alternative to GNSS-based navigation. However, existing vision-based techniques are inadequate in addressing flight deviation caused by environmental disturbances and inaccurate position predictions in practical settings. 
In this paper, we present a novel angle robustness navigation paradigm to deal with flight deviation in point-to-point navigation tasks. Additionally, we propose a model that includes the Adaptive Feature Enhance Module, Cross-knowledge Attention-guided Module and Robust Task-oriented Head Module to accurately predict direction angles for high-precision navigation. To evaluate the vision-based navigation methods, we collect a new dataset termed as UAV\_AR368. Furthermore, we design the Simulation Flight Testing Instrument (SFTI) using Google Earth to simulate different flight environments, thereby reducing the expenses associated with real flight testing. Experiment results demonstrate that the proposed model outperforms the state-of-the-art by achieving improvements of 26.0\% and 45.6\% in the success rate of arrival under ideal and disturbed circumstances, respectively.
\end{abstract}

\section{Introduction}
The Unmanned Aerial Vehicle (UAV) is a remotely controlled unmanned aircraft equipped with sensors and Global Navigation Satellite System (GNSS).
Typically, UAVs rely on GNSS for precise positioning and continuous adjustment of flight angles towards the end point. However, various factors, such as environmental conditions (e.g., buildings, and mountains), weather conditions (e.g., rain, snow, fog), and magnetic interference often result in UAVs receiving weak or no signals from GNSS. Recently, vision-based navigation has emerged as a promising alternative to GNSS-dependent navigation, thanks to the constant availability of scene images and affordable image acquisition equipment.

The point-to-point navigation is a typical method in vision-based navigation, serving crucial applications such as cargo transportation, package delivery, and crop cultivation. 
Existing point-to-point navigation can be grouped into: \textit{classification-based} and \textit{matching-based methods}.

\begin{figure*}[!t]
    \centering
    \includegraphics[width=\linewidth]{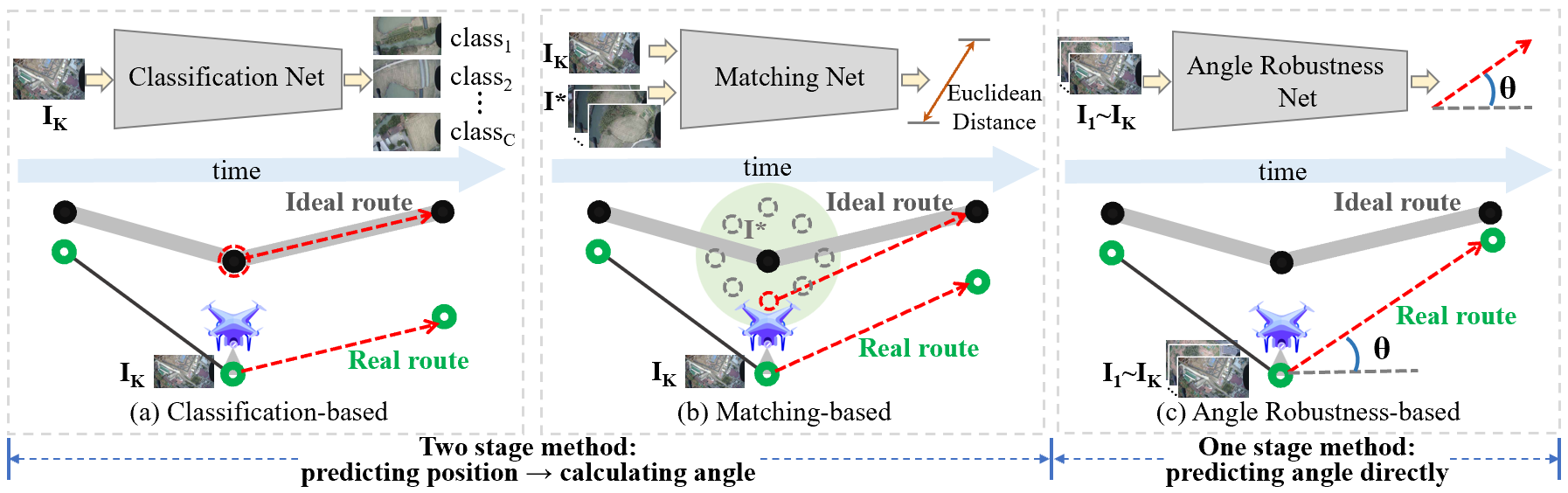}
    \caption{
    Comparison between the existing methods (a) Classification-based, (b) Matching-based methods and (c) Ours. (a) and (b) involve predicting the position and then calculating the direction angle. The inaccurate position prediction will lead to cumulative flight errors. In contrast, our method (c) directly predicts the direction angle and effectively solves flight deviation.}
    \label{fig:differ}
\end{figure*}

The former classifies the acquired image $I_K$ into a category with a specific position. Then, the direction guidance is computed by the category position and the ideal end point. However, flight deviations commonly occur as a result of wind and other disturbances. As shown in~\Cref{fig:differ} (a), the UAV in the deviated position flies with the direction guidance (denoted by the red dash line with an arrowhead). Following this guidance can result in cumulative position errors or even a wrong flight route.
The latter matches the acquired image $I_K$ in the deviated position with a set of candidate images $\{I^{\ast}\}$ within a specific area near the ideal position. Then, the direction guidance is computed by the matched position and the ideal end point. 
Although the matching-based approach has improved accuracy compared to classification-based method, it still has limitations including the requirements for significant storage space and time-consuming matching operations. Moreover, \Cref{fig:differ} (b) shows that incorrect matching will lead to a wrong flight direction (denoted by the red dash line with an arrowhead).
In sum, the aforementioned techniques are two-stage methods that firstly predict the position and then calculate the direction angle. Due to the disturbances in realistic scenarios, both of them are not efficient in solving flight deviation.

In this paper, we propose a novel angle robustness navigation paradigm, aimed at effectively addressing flight deviation caused by disturbances in realistic environments. As depicted in~\Cref{fig:differ} (c), in contrast to the traditional two-stage methods, the proposed paradigm directly predicts the angle based on historical frames, in order to adaptively find the correct direction. Specifically, an angle robustness navigation model that incorporates several key modules is put forward. The Adaptive Feature Enhance Module extracts sequential visual features from previous images, while the Cross-knowledge Attention-guided Module integrates path semantic information within the embedding sequences. To enhance robustness against environmental disturbances, Robust Activation Module is presented as an additional component of the Robust Task-oriented Head Module.

Furthermore, we collect a new dataset termed as UAV\_AR368 and design the Simulation Flight Testing Instrument (SFTI) using Google Earth. The SFTI is a testing tool capable of simulating diverse flight scenarios with disturbance factors like random wind, cutout, rain, and bright variation. By leveraging SFTI, we can efficiently evaluate the vision-based point-to-point navigation methods in real-world settings in a cost-effective manner.
Experiment results show that the proposed method significantly improves the success rate of arrival by 26.0\% and 45.6\% under the ideal and disturbed circumstances respectively.

Our contribution is therefore the first angle robustness navigation paradigm and model against flight deviation in realistic scenarios. We depart from the conventional two-stage navigation methods and adopt a more robust approach of directly predicting the direction angle. To enable efficient evaluation of vision-based navigation methods, a new dataset named UAV\_AR368 and a flight testing tool called SFTI are introduced. Extensive experiment results demonstrate that the proposed method achieves success rate improvements of 26.0\% and 45.6\% under ideal and disturbed circumstances respectively, compared to the state-of-the-art.

\section{Related Work}
Existing vision-based UAV navigation methods can be categorized into four main types: \textit{obstacle detection and avoidance}, \textit{map-independent navigation}, \textit{map-building navigation} and \textit{point-to-point navigation}. 
The first one~\cite{avoid2, avoid3, avoid6, avoid4, avoid1, adadrone} primarily concentrates on detecting and avoiding obstacles along flight routes. 
In this paper, we focus on the latter three techniques most related to our work.

\textbf{Map-independent Navigation.} 
This method navigates UAVs only by extracting features from surrounding circumstances instead of relying on maps. It mainly serves indoor corridor navigation~\cite{deepneural, dcnnga} and outdoor autonomous driving navigation~\cite{towards, interpretable}. \citet{endtoend} incorporated a novel FCN-LSTM architecture leveraging scene segmentation side tasks to improve autonomous driving performance. Unlike map-independent navigation, our approach focuses on navigating the UAV to the specific end point by learning knowledge from maps.

\textbf{Map-building Navigation.} 
In complex navigation environments, generating maps proves to be an effective solution. Map-building systems have been widely used in autonomous and semi-autonomous fields with the rapid development of visual simultaneous localization and mapping (SLAM) technology~\cite{orbslam, sptam}. A 3D LiDAR SLAM was presented by~\citet{slam1} for the localization of autonomous vehicles through encoding unordered point clouds with various resolutions. To reduce the influence of dynamic objects, \citet{slam2} introduced SIIS-SLAM based on sequential image segmentation integrating the optical flow dynamic detection module. Although the SLAM algorithm is mature, its steps in practical applications are relatively complex. In addition, the performance of SLAM tends to diminish in outdoor navigation scenarios with drastic environmental changes, compared to indoor settings.

\begin{figure*}[!t]
  \centering
  \includegraphics[width=\linewidth]{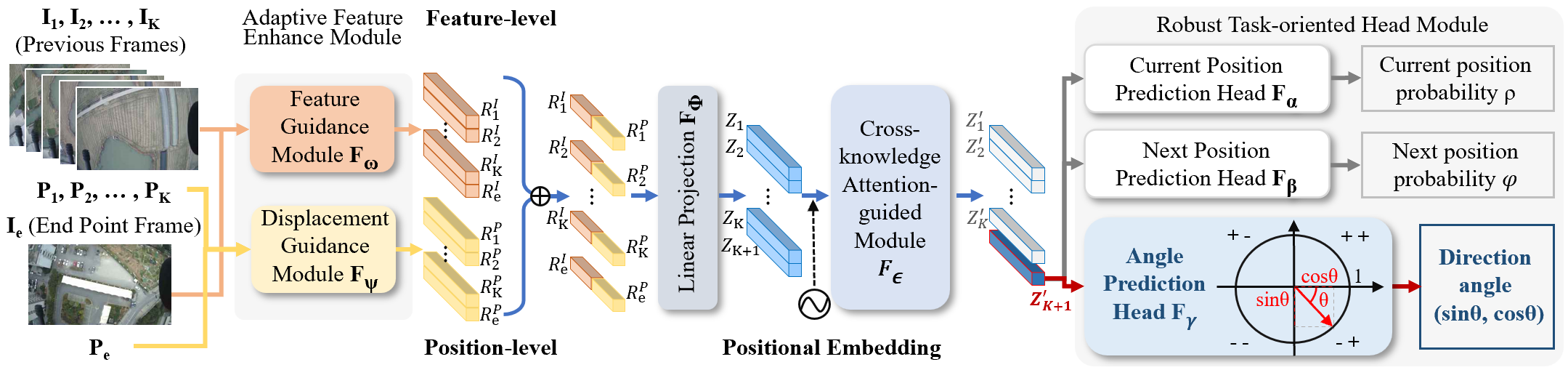}
  \caption{Architecture overview of the proposed method. Firstly, the Adaptive Feature Enhance Module extracts both sequential feature-level and position-level features $R^{I}$, $R^{P}$. Then, the Cross-knowledge Attention-guided Module integrates these semantic features to a global perspective embedding $Z'_{K+1}$. Finally, the Robust Task-oriented Head Module utilizes the embedding to adaptively predict the direction angle $\theta$.}
  \label{fig:armodel}
\end{figure*}

\textbf{Point-to-point Navigation.} 
This approach firstly determines the position of the UAV and then calculates the flight angle towards the intended end point.
\citet{university} introduced a new multi-view multi-source benchmark for UAV-based geolocalization task, termed as University-1652. This seminal work has spawned several impressive works. 
\citet{llfe} designed dense local feature descriptors for improving positioning accuracy. \citet{multiple-environment} and \citet{asemantic} proposed a dual branch framework against the problem of real-time images with extreme appearance changes. Apart from classification-based techniques, matching-based systems have also been employed in certain studies. To make full use of difficult positive samples, \citet{fine-grained} presented fine-grained image-region similarity. 
\citet{eachpart} proposed a square-ring feature partition strategy to extract key features. In order to avoid the fine-grained information lost, \citet{fsra} designed the Feature Segmentation and Region Alignment (FSRA) to enhance the model’s ability to understand contextual information. \citet{rknet} introduced RK-Net to learn a viewpoint-invariant visual representation.

Although the classification-based method is fast and simple, it is susceptible to serious flight deviation due to a single positioning error. Upon this, the matching-based method employs finer-grained matching operations to improve accuracy, but it requires substantial storage space and time. Moreover, it still suffers from flight deviation issues. In sum, the above methods are unable to address flight deviation in realistic environments.

\section{Task Definition}
In this section, the angle robustness point-to-point navigation paradigm is defined formally.
Given the previous frames $\mathcal{X} = (X_1, X_2, \ldots, X_K)$ captured by the UAV camera, where $K$ is the number of frames, the task is to train a network that predicts a direction angle $\theta$ within the range of $(-180^{\circ}, 180^{\circ}]$. This angle serves as the navigation command.
Unlike traditional two-stage approaches that firstly predict the position and then calculate the direction angle based on the predefined end point as depicted in~\Cref{fig:differ} (a) and (b), the angle robustness paradigm directly outputs the direction angle in an end-to-end manner thereby addressing flight deviation efficiently.

\section{Method}
Now we describe the proposed angle robustness point-to-point navigation model in detail.
Our method takes a sequence of $K$ continuous frames $\mathcal{X} = (X_1, X_2, \ldots, X_K)$ along with an end point frame $X_e$ as input ($K \geq 1$). It predicts a direction angle $\theta$ as the navigation command. Here, $X_K$ refers to the frame at the current timestamp, while $(X_1, X_2,\ldots, X_{K-1})$ indicate the previous historical frames.
Particularly, a frame $X_k \in \mathcal{X}$ is a tuple $(I_k, P_k)$ composed of a captured image $I_k$ and corresponding coordinate \begin{math}
    P_k\in\mathbb{R}^2
\end{math}.

\Cref{fig:armodel} shows the overall framework of the proposed method.
To begin with, the Adaptive Feature Enhance Module represents the image and position sequence as two embeddings $R^{I}$ and $R^{P}$, separately by the Feature Guidance Module and Displacement Guidance Module.
Then, the Transformer-based~\cite{transformer} Cross-knowledge Attention-guided Module merges local sequential features to a global perspective embedding $Z'_{K+1}$, which will be sent to the subsequent Robust Task-oriented Head Module to predict the direction angle $\theta$ adaptively. In order to improve the robustness of the model against disturbed situations, Robust Activation Module is incorporated into the Angle Prediction Head.
Furthermore, two auxiliary heads are introduced to facilitate the model's understanding of complex scenarios.

\subsection{Adaptive Feature Enhance Module}
The Adaptive Feature Enhance Module is proposed to extract sequential visual semantic information from the input $\mathcal{X}$ in a computational efficient manner. 
To achieve this, it introduces a Feature Guidance Module and a Displacement Guidance Module to extract key features.

\textbf{Feature Guidance Module.} We follow~\cite{mobilenetv3} and employ MobileNetV3-Small pre-trained on ImageNet~\cite{imagenet} as the backbone to extract valuable information.
Specifically, given a frame 
\begin{math}
    X_k=(I_k, P_k), (1 \leq k \leq K \mbox{ or } k = e)
\end{math}, the Feature Guidance Module denoted as $F_\omega(\cdot)$ with parameter set $\omega$, takes the image $I_k$ as the input and generates a feature embedding $R^{I}_k$ from the global average pooling layer:
\begin{equation}
R^I_k = F_\omega(I_k).
\end{equation}

\textbf{Displacement Guidance Module.} Apart from the feature-level information, the position-level information along the route is further provided by the Displacement Guidance Module so that the navigation model is able to generate more robust and accurate navigation commands.
Formally, for the k-th frame $X_k = (I_k, P_k)$, the displacement is defined as
\begin{equation}
\Delta P_{k}=\left\{\begin{array}{ll}
\frac{P_{k+1}-P_{k}}{\left\|P_{k+1}-P_{k}\right\|_{2}}, & \text { if } 1 \leq k<K, \\
(0,0), & \text { if } k=K \text { or } e.
\end{array}\right.
\end{equation}

Next, the Displacement Guidance Module denoted as $F_\psi(\cdot)$ with parameter set $\psi$, performs a shift add and a linear projection to map the displacement vector $\Delta P_k$ to an embedding $R^P_k$:
\begin{equation}
  \Delta P_k += \Delta P_{k - 1}, 1 < k \leq K \mbox{ or } e,
\end{equation}
\begin{equation}
  R^P_k = F_\psi(\Delta P_k).
\end{equation}

With the two components of the Adaptive Feature Enhance Module, the model input is mapped into a token sequence $\mathcal{Z}$ as follows: 

\begin{equation}
Z_k = F_{\phi}(R^I_k \oplus R^P_k),
\end{equation}
\begin{equation}
\mathcal{Z}=[Z_1, Z_2, \ldots, Z_k, \ldots, Z_K, Z_{K+1}],
\end{equation}
where $\oplus$ represents the concatenation operation, and $F_{\phi}(\cdot)$ denotes a linear projection with parameter set $\phi$. 
Through the concatenation and projection operations, position offset and image rotation information are incorporated into the feature embedding $Z_{k}$, which greatly promotes the model's ability to extract essential features from realistic images.

\subsection{Cross-knowledge Attention-guided Module}
In order to capture the interactive features present in the historical frame sequence, the Cross-knowledge Attention-guided Module based on Transformer is put forward.
Inspired by~\citet{gpt}, the Cross-knowledge Attention-guided Module is designed as a decoder-only Transformer. To be specific, the module consists of a learnable Positional Embedding layer, a Dropout layer and $L$ identical decoder layers. Each decoder layer is composed of a Masked Multi-Head Attention layer and a Feed Forward Network~(FFN).

Let $F_\epsilon(\cdot)$ denotes the Cross-knowledge Attention-guided Module with parameter set $\epsilon$. The output $\mathcal{Z}'$ is obtained by applying this module as follows:
\begin{equation}
 \mathcal{Z}'=F_\epsilon(\mathcal{Z}).
\end{equation}
Similar to $\mathcal{Z}$, $\mathcal{Z}'$ is composed of $K+1$ tokens denoted as $\{Z'_k\}^{K+1}_{k=1}$.
Instead of preserving the entire sequence, only the $(K+1)$-th token $Z'_{K+1}$ which contains sufficient image and angle features is adopted to predict the direction angle.

\subsection{Robust Task-oriented Head Module}

\textbf{Angle Prediction Head.} This head is presented to predict the direction angle adaptively. A LayerNorm layer and a FFN layer constitute its backbone. To improve the robustness of the model against various environmental disturbances, a Robust Activation Module is incorporated between the LayerNorm layer and FFN layer additionally. The Robust Activation Module is composed of a HardTanh activation function, a FFN layer and another HardTanh layer, where the FFN layer reduces the feature dimension of $Z'_{K + 1}$ to half of its original dimension.

Formally, the Angle Prediction Head denoted as $F_\gamma(\cdot)$ with parameter set $\gamma$, takes $Z'_{K+1}$ as input and generates a vector $(\sin{\theta}, \cos{\theta})$. Here, $\theta$ refers to the direction angle in degrees within the range of $(-180^{\circ}, 180^{\circ}]$. The direction angle is represented in the format of a vector including sine and cosine values for the following reasons:

Firstly, since the concerned task is point-to-point navigation, the direction from the start point to the end point remains fixed. This representation strategy divides the plane into four fine-grained quadrants thereby making it easier to determine the overall direction and decreasing the possibility of serious flight deviation caused by one-time prediction error.
Secondly, this strategy saves search space. As once the overall direction is determined, the other three quadrants can be disregarded.
Finally, predicting sine and cosine values simultaneously facilitates model learning. If the output angle is simply predicted within the range of $[-1, 1]$ and then linearly converted into the range of $(-180^{\circ}, 180^{\circ}]$, there is a risk that the angle may exceed the allowable range. In this case, the angle can only be rigidly truncated, which impairs the model's learning of nonlinear relationships. In contrast, the angle in degrees is calculated based on the arctangent value in our method:
\begin{equation}
\theta=\operatorname{arctan}(\frac{\sin{\theta}}{\cos{\theta}}) \div \pi \times 180^{\circ},
\end{equation}
and then converted into the correct range:
\begin{equation}
\theta=
\begin{cases}
\theta + 180^{\circ}, \mbox{ if } \sin{\theta} \geq 0 \mbox{ and } \cos{\theta} < 0, \\
\theta - 180^{\circ}, \mbox{ if } \sin{\theta} < 0 \mbox{ and } \cos{\theta} < 0.
\end{cases}
\end{equation}
Upon this, even if the output sine and cosine values exceed the range of $[-1, 1]$, the arctangent value is still within the allowable range, ensuring the degree value of the direction angle is always reasonable. Consequently, the representation strategy avoids truncation defects and promotes the model's learning of nonlinear relationships.

To enhance the cognitive capability of the model in complex scenarios, two auxiliary heads have been designed including the Current and the Next Position Prediction Head.

\textbf{Current Position Prediction Head.} The Robust Activation Module is omitted from this head compared to the Angle Prediction Head. Formally, given a label space with C categories in the dataset denoted as 
\begin{math}
\mathbb{C} = \left\{class_{1}, class_{2}, \ldots, class_{C}\right\}
\end{math}, 
the Current Position Prediction Head $F_{\alpha}(\cdot)$ with parameter set $\alpha$ maps the feature $Z'_{K+1}$ to a C-dimensional embedding $\rho \in \mathbb{R}^{C}$, which represents the class probabilities of the current position of UAV.

\textbf{Next Position Prediction Head.} This head has the same components as the above head. Differently, the Next Position Prediction Head $F_{\beta}(\cdot)$ with parameter set $\beta$, predicts the class probabilities $\varphi \in \mathbb{R}^{C}$ of the next position of UAV.

\begin{figure*}[!t]
    \centering
    \includegraphics[width=\linewidth]{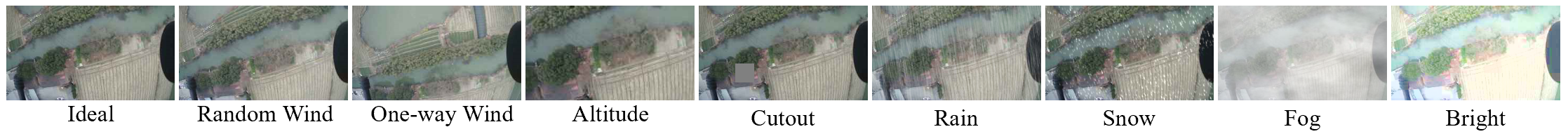}
    \caption{Samples of simulated flight testing environments.
    }
    \label{fig:disturbance}
\end{figure*}

\section{Simulation Flight Testing Instrument}
\subsection{Validation Process}
During testing, the coordinates of the images along the route cannot be obtained from GNSS, except for those of the start and the end points. In other words, at the beginning, only two frames are fed into the model: the start point $X_1=(I_1, P_1)$ and end point $X_e=(I_e, P_e)$. Then, the model predicts the direction angle $(\sin{\theta_1}, \cos{\theta_1})$ and calculates the next position $\hat{P_2}$: 
\begin{math}
\hat{P_2} = P_1 + (\sin{\theta_1}, \cos{\theta_1}) \times D,
\end{math} where D is the UAV flight distance each flight step. Following the procedure depicted in the next subsection, $(X_1, X_2, X_e)$ is obtained as the model input of the next prediction step. Apparently, each prediction step increases the input length by 1. The above all processes are repeated until the input length (excluding $X_e$) exceeds $K$, then the earliest frame is discarded to maintain a constant input length. The testing route is considered complete when the UAV successfully reaches the predefined end point within 25 m / 50 m, or the number of flight steps exceeds 250, or the UAV flies out of the designated map range of $5328 \times 2300 m^2$:
\begin{equation}
\left\|P_e - \hat{P_K}\right\|_{2} \leq 25 \mbox{ or } \left\|P_e - \hat{P_K}\right\|_{2} \leq 50,
\end{equation} where $P_e$ indicates the ideal end point position, and $\hat{P_K}$ refers to the predicted position at the current timestamp.

\subsection{Setting of SFTI}
In order to achieve efficient evaluation of vision-based navigation methods in real-world settings, we present the Simulation Flight Testing Instrument (SFTI) using Google Earth as its satellite images are very similar to realistic UAV images. It is assumed that the UAV is flying parallel to the ground, with a pitch of $90^{\circ}$ and a flight distance of 30 m each flight step. The ideal flight altitude is 80 m. The counterclockwise rotation angle from the eastward direction line is positive, and the clockwise rotation angle is negative. The role of the SFTI in each prediction step is as follows: 
Firstly, the direction angle is predicted and the coordinate of the next position is calculated. Secondly, since there are cases that the UAV needs to adjust the flight parameters or encounters wind, the Horizontal Position Deviation (Random Wind, One-way Wind) and / or Vertical Position Deviation (Altitude) are added to the original position. Thirdly, the screenshot is captured on Google Earth based on the adjusted position. Fourthly, imgaug~\cite{imgaug} library is utilized to process the screenshot to imitate random environmental disturbances, including Environmental Conditions (Cutout), Weather Conditions (Rain, Snow, Fog) and Lighting Conditions (Bright). Finally, the model input for the next prediction step is obtained. ~\Cref{fig:disturbance} shows the examples of the ideal and disturbed testing environments.

\begin{table*}[!t]
\small
\resizebox{\linewidth}{!}{
  \begin{tabular}{cccccccccccccccc}
    \toprule
     \multirow{2}{*}{Type} & \multirow{2}{*}{Method} & \multicolumn{2}{c}{SR@25 $\uparrow$} & \multicolumn{2}{c}{SR@50 $\uparrow$} & \multicolumn{2}{c}{MEPE $\downarrow$} & \multicolumn{2}{c}{MRE $\downarrow$} & \multicolumn{2}{c}{MRD $\uparrow$} & \multirow{2}{*}{\makecell{Inference \\Time}} & \multirow{2}{*}{Storage}\\
     
     \cmidrule(r){3-4} \cmidrule(r){5-6} \cmidrule(r){7-8} \cmidrule(r){9-10} \cmidrule(r){11-12}
     
     & & w/o & w/ & w/o & w/ & w/o & w/ & w/o & w/ & w/o & w/ & &\\

    \midrule

    \multirow{3}{*}{Classify} & MobileNetV3 & 0.0\% & 0.0\%  & 0.0\% & 0.0\% & - & - & - & - & - & - & 9.0 & 0.0\\
    
    & ResNet18 & 0.0\% & 0.0\%  & 0.0\% & 0.0\% & - & - & - & - & - & - & 58.0 & 0.0\\
    
    & ViT-B/16 & 0.0\% & 0.0\%  & 0.0\% & 0.0\% & - & - & - & - & - & - & 102.0 & 0.0\\

    \cmidrule(r){1-2} \cmidrule(r){3-4} \cmidrule(r){5-6} \cmidrule(r){7-8} \cmidrule(r){9-10} \cmidrule(r){11-12} \cmidrule(r){13-14} 
    
    \multirow{3}{*}{Match} & FSRA & 3.0\% & 1.9\% & 7.0\% & 3.8\% & 28.3 & 24.2 & 111 & 109 & 232 & 207 & 265.0 & 247.9\\

    & RK-Net & 1.0\% & 0.6\% & 3.0\% & 1.9\% & 33.4 & 38.8 & 108 & 113 & 420 & 291 & 203.0 & 122.6\\

    & LPN & 3.0\% & 1.3\% & 4.0\% & 3.8\% & 20.1 & 33.5 & 106 & 107 & 171 & 172 & 174.0 & 166.3\\

    \cmidrule(r){1-2} \cmidrule(r){3-4} \cmidrule(r){5-6} \cmidrule(r){7-8} \cmidrule(r){9-10} \cmidrule(r){11-12} \cmidrule(r){13-14} 

    \multirow{3}{*}{\makecell{Angle \\Robust}} & Backbone & 6.0\% & 3.1\% & 8.0\% & 3.8\% & 21.9 & \textbf{16.7} & 59 & 61 & 579 & 618 & 40.0 & 0.0\\

    & Backbone-FC & 74.0\% & 21.9\% & 74.0\% & 35.6\% & 15.3 & 26.8 & 30 & 41 & 3783 & 2288 & 41.0 & 0.0\\
    
    & Ours & \textbf{100.0\%} & \textbf{67.5\%} & \textbf{100.0\%} & \textbf{78.8\%}& \textbf{13.2} & 19.7 & \textbf{20} & \textbf{31} & \textbf{4484} & \textbf{3820} & 49.0 & 0.0\\

    \bottomrule
  \end{tabular}
}

\caption{
  Experiment results w/o and w/ disturbances on UAV\_AR368. SR@25, SR@50 (The Success Rate of arrival within 25 m and 50 m from the end point), MEPE (The Mean End Point Error on the premise of successful arrival, with the unit of meters), MRE and MRD (The Mean Route Error and the Mean Route Distance on the premise of not deviating from the prescribed route, with the unit of meters), as well as Inference Time (ms) and Storage (MB) are reported. $\uparrow$ represents that larger values are better. $\downarrow$ represents the opposite. When the SR@25 and SR@50 are 0.0\%, the measurements of MEPE, MRE, and MRD are not possible, as indicated by the symbol ``-". Bold indicates the best results.
  }
\label{tab:results}

\end{table*}

\begin{figure*}[!t]
    \centering
    \includegraphics[width=\linewidth]{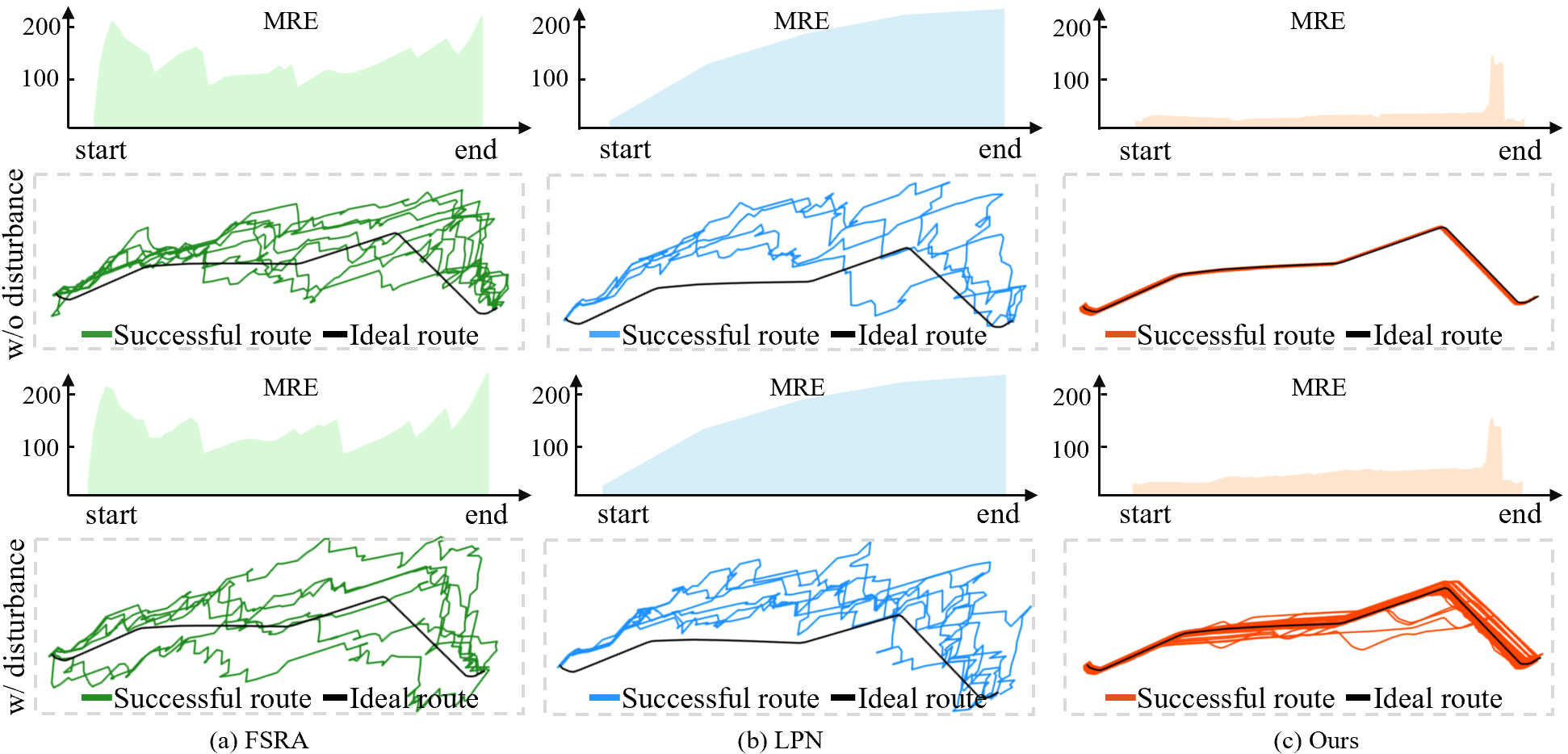}
    \caption{Visualization results w/o and w/ disturbances on UAV\_AR368. The subgraphs in the coordinate systems above indicate the MRE along the testing routes. The gray dash-line boxes below illustrate the flight routes generated by different methods, where the successful and ideal routes are denoted by green / blue / red and black lines.}
    \label{fig:visualtest}
\end{figure*}

\section{Experiments}

\subsection{Experiment Settings}
\textbf{Datasets.} We collect a new dataset termed as UAV\_AR368 containing 368 routes and 56,880 images on the topic of UAV point-to-point navigation. Specifically, the UAV adopts a speed range of $[5, 15]$ m / s and flies at an altitude range of $[60, 100]$ m to capture 184 routes to cope with situations where flight parameters need to be adjusted during testing. Each route is over 5,800 m. Each image has a resolution of $320 \times 180$ and is accompanied by corresponding GNSS coordinate. Due to the challenges of capturing images under extreme weather conditions, imgaug library is employed to process these images and generate additional 184 routes, including Cutout, Rain, Snow, Fog, and Bright. In sum, UAR\_AR368 is a multi-domain and multi-scale dataset with various environmental conditions and field of view size. Finally, the dataset is classified into 100 categories based on geographical positions and split into a training set with 45,398 images and a testing set with 11,482 images.

To our knowledge, there are no publicly available datasets for UAV point-to-point navigation tasks. Existing datasets related to UAV missions are aimed at solving UAV localization problems, and composed of discrete regions, such as University-1652~\cite{university} and CVUSA~\cite{cvusa}. In contrast, our dataset mainly focuses on the point-to-point navigation from the start point to the end point which requires continuous routes. This consideration has prompted us to propose our own dataset.

\textbf{Implementation Details.} 
The value of $K$ is 5 (excluding $X_e$). All input images are randomly cropped and resized to $224 \times 224$.
The dimensions of $R^I_k$, $R^P_k$, $Z_k$, $Z'_{k}$ are 576, 512, 512, 512. The dimension, depth, and hidden dimension of the Cross-knowledge Attention-guided Module are 512, 4, and 1024. An eight-head strategy is adopted in the Masked Multi-Head Attention layer. The dropout rate is 0.1.
The architecture is optimized by AdamW optimizer with a learning rate of 1e-3 and a weight decay of 0.1. The CosineAnnealingWarmRestarts is employed as the learning rate decay scheduler and Task-Uncertainty Loss~\cite{uncertainty} is used as the loss function. The model is trained for 120 epochs with the batch size of 128. We conduct 100 and 160 flight tests with a length of over 5,800 m for ideal and disturbed environments. D is set to 30 m.

\textbf{Baselines.} Eight vision-based methods are compared in the experiments: 1) MobileNetV3-Small~\cite{mobilenetv3}; 2) ResNet18~\cite{resnet}; 3) ViT-B/16~\cite{vit}; 4) FSRA~\cite{fsra}; 5) RK-Net~\cite{rknet}; 6) LPN~\cite{eachpart}; 7) Backbone, which removes the Displacement Guidance Module from the proposed architecture; 8) Backbone-FC, which removes the Robust Activation Module from the proposed architecture.

\begin{table*}[!t]
\small
\resizebox{\linewidth}{!}{
  \begin{tabular}{cccccccccccccc}
    \toprule
     \multirow{2}{*}{Method} & \multirow{2}{*}{\makecell{Candidate \\Number}} & \multicolumn{2}{c}{SR@25 $\uparrow$} & \multicolumn{2}{c}{SR@50 $\uparrow$} & \multicolumn{2}{c}{MEPE $\downarrow$} & \multicolumn{2}{c}{MRE $\downarrow$} & \multicolumn{2}{c}{MRD $\uparrow$} & \multirow{2}{*}{\makecell{Inference \\Time}} & \multirow{2}{*}{Storage}\\
     
     \cmidrule(r){3-4} \cmidrule(r){5-6} \cmidrule(r){7-8} \cmidrule(r){9-10} \cmidrule(r){11-12}
     
     & & w/o & w/ & w/o & w/ & w/o & w/ & w/o & w/ & w/o & w/ & &\\
     
    \midrule
    
    \multirow{3}{*}{FSRA} & 25 & 3.0\% & 0.6\% & 5.0\% & 2.5\% & \textbf{25.7} & 34.2 & \textbf{107.0} & 111.0 & \textbf{261.0} & \textbf{239.0} & 286.0 & 187.8\\

    & 33 & \textbf{3.0\%} & \textbf{1.9\%} & \textbf{7.0\%} & 3.8\% & 28.3 & \textbf{24.2} & 111.0 & \textbf{109.0} & 232.0 & 207.0 & 265.0 & 247.9\\

    & 41 & 2.0\% & 0.6\% & 6.0\% & \textbf{4.4\%} & 33.5 & 35.8 & \textbf{107.0} & 110.0 & 247.0 & 229.0 & 278.0 & 308.0\\

    \cmidrule(r){1-2} \cmidrule(r){3-4} \cmidrule(r){5-6} \cmidrule(r){7-8} \cmidrule(r){9-10} \cmidrule(r){11-12} \cmidrule(r){13-14} 
    
    \multirow{3}{*}{RK-Net} & 17 & 0.0\% & 0.0\% & 0.0\% & 0.0\% & - & - & - & - & - & - & 176.0 & 88.3\\

    & 25 & \textbf{1.0\%} & \textbf{0.6\%} & \textbf{3.0\%} & \textbf{1.9\%} & \textbf{33.4} & \textbf{38.8} & \textbf{108.0} & \textbf{113.0} & \textbf{420.0} & \textbf{291.0} & 203.0 & 122.6\\

    & 33 & 0.0\% & 0.0\% & 1.0\% & 1.3\% & 46.3 & 39.8 & 116.0 & 114.0 & 286.0 & 258.0 & 190.0 & 161.8\\

    \cmidrule(r){1-2} \cmidrule(r){3-4} \cmidrule(r){5-6} \cmidrule(r){7-8} \cmidrule(r){9-10} \cmidrule(r){11-12} \cmidrule(r){13-14} 
    
    \multirow{3}{*}{LPN} & 9 & 1.0\% & \textbf{1.3\%} & 3.0\% & \textbf{3.8\%} & 29.0 & 33.5 & \textbf{106.0} & 107.0 & 168.0 & 172.0 & 162.0 & 88.0\\

    & 17 & \textbf{3.0\%} & 0.0\% & \textbf{4.0\%} & 1.3\% & \textbf{20.1} & 44.5 & \textbf{106.0} & \textbf{106.0} & \textbf{171.0} & 168.0 & 174.0 & 166.3\\

    & 25 & 1.0\% & 0.6\% & 2.0\% & 2.5\% & 31.1 & \textbf{28.8} & \textbf{106.0} & \textbf{106.0} & 170.0 & \textbf{177.0} & 211.0 & 244.5\\

    \bottomrule
  \end{tabular}
}

\caption{Experiment results on the ablation study on the number of candidate images for matching-based methods. The meanings of the symbols $\uparrow$, $\downarrow$, and ``-" are consistent with those in~\Cref{tab:results}. Bold indicates the best results.}
\label{tab:candinum}
  
\end{table*}

\textbf{Evaluation Metrics.}
To compare the performances of vision-based approaches, seven evaluation metrics are employed in the experiments: SR@25, SR@50, MEPE, MRE, MRD, Inference Time and Storage. The first two metrics represent the Success Rate of arrival within 25 m and 50 m from the end point. The MEPE represents the Mean End Point Error on the premise of successful arrival. The MRE and MRD denote the Mean Route Error and Mean Route Distance on the premise of not deviating from the prescribed route, respectively. Furthermore, to illustrate the efficiency of vision-based methods, Inference Time and Storage are compared. The units of MEPE, MRE, MRD are meters (m), while Inference Time and Storage are measured in milliseconds (ms) and megabytes (MB). The maximum deviation range for MRE and MRD is set to 200 m. 

\subsection{Quantitative Comparison Results}
The experiments are conducted on the UAV\_AR368 dataset.
In~\Cref{tab:results}, classification-based methods fail all flights with success rate of 0.0\%. They incorrectly classify the initial images, leading to severe direction error. Although the matching-based methods improve the success rate to around 3.0\%, they are still unable to meet the requirements for high-precision navigation. In contrast, the proposed angle robustness method produces superior performance. Specifically, the proposed method achieves SR@25 of 100.0\% and 67.5\% in ideal and disturbed environments, which is 26.0\% and 45.6\% higher than that of the baselines. It also achieves SR@50 of 100.0\% and 78.8\% in ideal and disturbed situations, which is 26.0\% and 43.2\% higher than that of the baselines. Additionally, the MEPE of Ours is around 13.7\% lower than the MEPE of the baselines, in ideal condition.

To verify the robustness of distinct techniques during flight, MRE and MRD are reported. The MRE of Ours is less than that of the baselines (20 m vs. 30 m; 31 m vs. 41 m). Ours achieves an MRD that is nearly 1.2 and 1.7 times that of the baselines in ideal and disturbed situations. Moreover, the proposed architecture has only 10.7 M \#Params and 419.7 M FLOPs, which is friendly to the UAV with limited memory and computing power. The inference time of Ours is only 49 ms, and no additional space is required beyond the deployment space for the model. To sum up, the angle robustness approach enables more accurate, robust and efficient point-to-point navigation for UAVs.

\begin{table}[!t]
\small
\resizebox{\linewidth}{!}{
  \begin{tabular}{ccccccccccccc}
    \toprule
    AFEM & CAM & CPH & NPH & RAM & w/o & w/ \\
    
    \midrule
    
    \usym{2717} & \usym{1F5F8} & \usym{1F5F8} & \usym{1F5F8} & \usym{1F5F8} & 0.0\% & 0.0\%\\

    \usym{1F5F8} & \usym{2717} & \usym{1F5F8} & \usym{1F5F8} & \usym{1F5F8} & 0.0\% &  0.0\%\\

    \usym{1F5F8} & \usym{1F5F8} & \usym{2717} & \usym{1F5F8} & \usym{1F5F8} & 63.0\% & 32.5\%\\

    \usym{1F5F8} & \usym{1F5F8} & \usym{1F5F8} & \usym{2717} & \usym{1F5F8} & 68.0\% & 47.5\%\\
    
    \usym{1F5F8} & \usym{1F5F8} & \usym{2717} & \usym{2717} & \usym{1F5F8} & 0.0\% & 0.6\%\\

    \usym{1F5F8} & \usym{1F5F8} & \usym{1F5F8} & \usym{1F5F8} & \usym{2717} & 74.0\% & 21.9\%\\
    
    \usym{1F5F8} & \usym{1F5F8} & \usym{1F5F8} & \usym{1F5F8} & \usym{1F5F8} & \textbf{100.0\%} & \textbf{67.5\%} \\
    
    \bottomrule
  \end{tabular}
  }

   \caption{Ablation results on different model components. SR@25 is reported. Bold indicates the best results.
  }
  \label{tab:ablresults}
  
\end{table}

\subsection{Qualitative Visualization Results} 
\Cref{fig:visualtest} depicts the visual results w/o and w/ disturbances. We can see that the proposed method achieves lower MRE. In addition, the tested successful routes of Ours are closer to the ideal route compared to the matching-based baselines. In disturbed environments, even if the UAV deviates during flight, the proposed method can still effectively correct the subsequent directions. These findings demonstrate that the proposed model is able to predict more robust and accurate direction angle compared to previous approaches.

\subsection{Ablation Study}
Ablation experiments on different model components are conducted and SR@25 in ideal and disturbed environments is reported in~\Cref{tab:ablresults}. Six novel components are considered: 1) Adaptive Feature Enhance Module (AFEM); 2) Cross-knowledge Attention-guided Module (CAM); 3) Current Position Prediction Head (CPH); 4) Next Position Prediction Head (NPH); 5) Robust Activation Module (RAM). We can see that the removal of the AFEM and CAM results in the success rate of 0.0\%, which proves their significant impact on semantic feature extraction and incorporation. 
It can be observed that the removal of the CPH and NPH reduces the SR@25 to 63.0\% and 68.0\%, respectively, indicating their crucial roles in enhancing the model's comprehension ability in complex scenarios. If two heads are both deleted, the success rate directly drops to 0.0\%.
Moreover, the RAM with HardTanh activation performs well by facilitating a more stable training process. 

In addition, ~\Cref{tab:candinum} presents the experiment results on the ablation study on the number of candidate images for matching-based baselines. We can see that the best number of candidate images for FSRA and RK-Net approaches is 33 and 25, while the best number for LPN approach is 17 and 9 in ideal and disturbed environments, respectively. 

\subsection{Discussion and Future Work}
From~\Cref{tab:results}, it can be seen that the point-to-point navigation model still cannot achieve 100.0\% success rate in disturbed testing environments. Possible reasons include multiple aspects: 1) insufficient training samples in the dataset; 2) single kind of testing routes with limited ground truth direction angles in the dataset; 3) the model may be over-/under-fitting. We will explore the latent reasons in the future work. 

In this paper, we verify the proposed model on limited route samples. In fact, the model can be extended to a general pre-training navigation framework capable of achieving arbitrary flight navigation, within a country, state, or even the world in GNSS-denied scenarios. We will focus on exploring the extension of the proposed method in the future.
 
\section{Conclusion}
In this paper, we propose an angle robustness navigation paradigm to effectively address UAV flight deviation in realistic environments. Upon this, an angle robustness navigation model is designed to directly predict the direction angle. Moreover, we collect a new dataset named UAV\_AR368 and design the Simulation Flight Testing Instrument (SFTI) using Google Earth, which can evaluate the performance of various navigation methods. Experimental results demonstrate that the proposed paradigm and model are practical solutions for vision-based navigation in realistic scenarios.

\section{Acknowledgments}
This work is funded by National Natural Science Foundation of China (U20B2066), Zhejiang Province High-Level Talents Special Support Program ``Leading Talent of Technological Innovation of Ten-Thousands Talents Program" (No. 2022R52046), the Ningbo 2025 Science and Technology Innovation Major Project (2022Z072), and Ningbo Natural Science Foundation (2022J182).

\bibliography{aaai24}

\end{document}